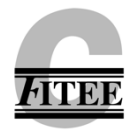

# FinBrain: When Finance Meets AI 2.0*

Xiao-lin ZHENG[†‡1], Meng-ying ZHU[1], Qi-bing LI[1], Chao-chao CHEN[2], Yan-chao TAN[1]

[1]Department of Computer Science, Zhejiang University, Hangzhou 310027, China
[2]AI Department, Ant Financial Services Group, Hangzhou 310027, China
[†]E-mail: xlzheng@zju.edu.cn

**Abstract:**    Artificial intelligence (AI) is the core technology of technological revolution and industrial transformation. As one of the new intelligent needs in the AI 2.0 era, financial intelligence has elicited much attention from the academia and industry. In our current dynamic capital market, financial intelligence demonstrates a fast and accurate machine learning capability to handle complex data and has gradually acquired the potential to become a "financial brain". In this work, we survey existing studies on financial intelligence. First, we describe the concept of financial intelligence and elaborate on its position in the financial technology field. Second, we introduce the development of financial intelligence and review state-of-the-art techniques in wealth management, risk management, financial security, financial consulting, and blockchain. Finally, we propose a research framework called FinBrain and summarize four open issues, namely, explainable financial agents and causality, perception and prediction under uncertainty, risk-sensitive and robust decision making, and multi-agent game and mechanism design. We believe that these research directions can lay the foundation for the development of AI 2.0 in the finance field.

**doi:**10.1631/FITEE.1000000           **Document code:**   A           **CLC number:**   TP39

## 1  Introduction

Artificial intelligence (AI), the core technology of new technological revolution and industrial transformation, is transcending the traditional means of simulating human intelligence by a computer, such as man–machine gaming, machine identification, and natural language processing. Consequently, a new generation of AI, namely, AI 2.0, has emerged. Pan defined AI 2.0 as a new generation of AI based on the novel information environment of major changes and the development of new goals (Pan 2016). As one of the new intelligent needs, financial intelligence is a highly suitable application setting for AI 2.0.

‡ Corresponding author
* Project supported by the National Natural Science Foundation of China (No.  U1509221)
 ORCID:  Xiao-lin ZHENG, http://orcid.org/0000-0001-5483-0366
© Zhejiang University and Springer-Verlag Berlin Heidelberg 2015

AI 2.0 as a core technology is an important driving force for the development and transformation of the financial industry. Technology-driven financial upgrading can be divided into three stages (Table 1). The first stage is Fintech 1.0. At this stage, computers can be used to replace manual calculation and accounting books to improve the efficiency of financial operations. The second stage is Fintech 2.0, which is also called Internet finance. At this stage, technology is a force for financial revolution and provides emerging Internet enterprises with the opportunity to use Internet technology to connect the supply and demand for financial products and services. Internet finance becomes an effective supplement to traditional finance. The third stage is Fintech 3.0, which is also known as intelligent finance or smart finance. At this stage, financial technology integrates the Internet, finance, and big data to achieve intelligent and accurate calculation responsibility and lead the overall change in the financial industry through big data, blockchain,

**Table 1 Main stages of technology-driven financial industry development**

| Development stage | Driving technology | Main landscape | Inclusive finance | Relationship between technology and finance |
|---|---|---|---|---|
| Fintech 1.0 (Financial IT) | Computer | Credit Card, ATM, CRMS | Low | Technology as a tool |
| Fintech 2.0 (Internet Finance) | Mobile Internet | Marketplace Lending, Third Party Payment, Crowdfunding, Internet Insurance | Medium | Technology-driven change |
| Fintech 3.0 (Financial Intelligence) | Big data, Blockchain, Cloud Computing, AI, etc. | Intelligent Finance | High | Deep fusion |

cloud computing, AI, and other emerging technologies.

Financial intelligence has a fast and accurate machine learning capability to achieve the intellectualization, standardization, and automation of large-scale business transactions. Thus, it can improve service efficiency and reduce costs. For example, AI technology combined with big data can integrate long-tail markets and mitigate information asymmetry to improve the efficiency of fund allocation and financial risk management. Moreover, identity recognition and natural language processing technology can allow machines to replace laborers, realize all-around perception, and provide interactive services to customers. By combining financial intelligence with the new generation of AI, we can construct a "financial brain" that offers inclusive financial services to meet the financial needs of ordinary people. This stage has three important characteristics.

**Rich usage scenarios.** Many financial applications, such as investment, lending, credit, security, insurance, and customer service, are inseparable from the support of AI technology.

**Highly structural business data.** Compared with other industries, the financial service industry produces large amounts of structural data and thus has the advantage of developing AI technology.

**Meets the requirements of inclusive finance.** Traditional professional financial services often have a high threshold; thus, the financial needs of ordinary people are not well met. The financial threshold can be significantly reduced by the effectiveness of credit assessment technology, and improved services can be provided to people. Hence, everyone can enjoy a fair opportunity to achieve inclusive finance.

The main purpose of this article is to explore the potential of constructing a "financial brain" in the forthcoming era of AI 2.0. To achieve this aim, we briefly review state-of-the-art advances in different areas, including wealth management, risk management, financial security, financial consulting, and blockchain. We propose a financial research framework called FinBrain and discuss several emerging open issues from the technical perspective to thoroughly understand challenges and future directions.

## 2 Research and Application

This section briefly reviews the recent progress in different areas of financial intelligence to demonstrate that AI technology can improve service efficiency and reduce costs.

### 2.1 Wealth management

The need to provide individuals with wealth-management services has emerged recently. Examples of such services include Robo-Advisor, financial product recommendation, and precision marketing that can help maintain good customer experience while mitigating risks and improving individuals' decision-making capabilities. Most wealth managers use simple rule-based analytics based on reporting systems that cannot effectively characterize customers' preferences (Bahrammirzaee 2010). The new generation of AI 2.0 shows a remarkable potential to provide enhanced wealth-management services via user profiling, predictive analytics (Ding, Zhang et al. 2015, Flunkert, Salinas et al. 2017), the Internet of Things (IoT) intelligence (Dineshreddy and Gangadharan 2016), and customized recommendations (Zhao, Wu et al. 2014, Zhang, Yao et al. 2017). For example, smart agents can combine structured financial and unstructured behavioral data to assess customers' investment style, risk tolerance, and purchasing preferences and develop optimized portfolios at low costs and high speeds. Furthermore, firms can utilize information from clients' IoT ecosystems to tailor investment decisions and asset allocations based on their individual behaviors, preferences, and locations. We describe below the recent development of two key wealth-management applications, namely, financial product recommendation and Robo-Advisor.

**Financial Product Recommendation:** The key to designing a good recommendation system is to characterize the preferences of customers and construct personalized behavior models. Conventional recommendation approaches are frequently combined with Markowitz's portfolio theory to achieve improved investment recommendations (Zhao, Wu et al. 2014, Zhao, Liu et al. 2016). For example, Zhao, Wu et al. (2014) constructed a two-step decision making platform for investment recommendation that considers what to buy and how much money to pay. Zhao, Liu et al. (2016) were the first to assess P2P loans from a multi-objective perspective and recommend portfolios, such as non-default and winning-bid probabilities. They combined the static and dynamic features of bidding lenders into a gradient boosting decision tree (GBDT) framework. Recent state-of-the-art recommendation models adopt deep neural networks (DNNs) to effectively model highly complex rating functions (He, Liao et al. 2017) or exact high-order feature interactions (Guo, Tang et al. 2017), and they may open a new avenue of financial product recommendation based on deep learning.

**Robo-Advisor:** The core task of Robo-Advisor is to build an optimized portfolio for asset management by using risk models, which help decrease the cost of portfolio construction while improving quality. Several researchers have incorporated multi-armed bandits into the sequential decision-making process to achieve improved portfolio blending (Wang, Wang et al. 2015, Shen and Wang 2016), in which online portfolio choice algorithms are developed, and the tradeoff between exploration and exploitation is further considered. Jiang, Xu et al. (2017) presented a financial reinforcement learning framework based on deterministic policy gradient to allocate portfolios continuously and maximize the investment return. In their work, an environment state was represented by a price tensor that summed up the historic price data in the market, and the agent's action was represented a continuous portfolio weight vector. Recently, Heaton, Polson et al. (2016) employed deep learning models (e.g., auto-encoders and long short-term memory units) to form hierarchical decision models for portfolio construction and risk management that can reconstruct invisible non-linear interactions. The authors proposed a deep portfolio theory that involves a four-step routine, namely, encode, calibrate, validate, and verify, to formulate an automated and general portfolio selection process. Predictive analysis (Ding, Zhang et al. 2015, Flunkert, Salinas et al. 2017) is also critical when providing Robo-Advisor services because it offers an enhanced understanding of market trends and helps identify investment opportunities.

## 2.2 Risk Management

Risk management includes the identification, measurement, and control of financial risks. The traditional risk management process is highly complicated and relies on the experience of experts. However, with the combination of smart phones and E-commerce, consumers are allowed to interact with merchants conveniently. The Internet is transitioning to the IoT. Thus, traditional risk management cannot meet the financial needs of most people due to their high time and manpower costs and low coverage. Meanwhile, AI technology combined with big data can help build robust credit systems, assess business risks under uncertainty, and achieve anti-fraud functions. For example, diversified data sources can be utilized to build a credit system based on the tree model (Yeh, Lin et al. 2012), neural network (Khashman 2010, Abdou, Tsafack et al. 2016), and support vector machine model (Han, Han et al. 2013). Moreover, businesses rely increasingly on AI technology for competitive advantages, and IoT provides risk managers abundant information for assessing and managing risks. For example, by using IoT data, such as mobile, household devices and other sensing devices, banks can analyze geographical data and identify the real trading environments, so that they can alert the customer before someone swipes their credit card in a mall (Dineshreddy and Gangadharan 2016). Hence, operations do not have to be reinvented. This feature provides organizations that are reliant on managing risks with an indispensable tool. The following text discusses the four latest key applications of risk management, namely, intelligent credit, risk assessment, fraud detection, and bankruptcy prediction.

**Intelligent credit:** Credit evaluation is the basis of risk management. Traditional financial credit information usually considers only strong financial attributes, such as credit, credit card, foreign exchange, private lending, and other financial transaction data. For example, FICO selects indicators, such as length of residence, job, length of service, debt-to-income ratio, and line of credit. Subsequently, it calculates the weighted average as the final score on the basis of expertise-based rules. Different from traditional credit information systems, intelligent credit systems, which integrate big data and AI, consider financial, government public service, life, and social data (Angelini, Tollo et al. 2008). These systems break the data island and cover diversified data sources (Bahrammirzaee 2010), and their corresponding models are more complex than traditional credit models (Hoofnagle 2014). In addition, deep learning (Ha and Nguyen 2016) is an effective method to manage high-dimensional credit characteristic data. ZestFinance is a credit evaluation system based on machine learning. It accesses large amounts of data from thousands of data items and tens of thousands of variables, establishes numerous credit risk prediction sub models, and integrates these models into learning to obtain comprehensive credit scores. Zhima Credit combines the models of logistic regression, neural network, and decision tree and considers five aspects: personal identity, interpersonal relationship, credit record history, behavioral preference, and individual performance ability.

**Table 2 Comparison of three personal credit rating systems**

|  | Traditional credit evaluation | Credit evaluation based on machine learning |  |
|---|---|---|---|
| Representative enterprise | FICO | ZestFinance | Representative enterprise |
| Service user | Rich credit records | Lack or absence of credit records | Service user |
| Data content | Financial data | Financial data, life data | Data content |
| Data sources | Bank | Data from third parties and data provided by users | Data sources |
| Model | Logistic regression | Machine learning | Model |

**Risk assessment:** Compared with traditional approaches that rely heavily on the availability of rich financial information and the experience of experts, AI technology can automatically identify hidden patterns through heterogeneous data sources and can thus achieve improved user profiling when assessing risks. For example, Node2vec (Grover and Leskovec 2016) and Struc2vec (Ribeiro, Saverese et al. 2017) can effectively learn the latent representations of a node's structural identity. Risk assessment systems can further integrate the learnt latent representations into GBDT or DNNs to understand the relational network that involves merchants, consumers, investors, and borrowers. Zhou, Ding et al. (2017) designed KunPeng, a universal distributed platform for large-scale machine learning. KunPeng provides risk estimation for Alibaba's Double 11 Online Shopping Festival and Ant Financial's transactions. Khandani, Kim et al. (2010) constructed a machine-learning forecasting model to estimate consumer credit risks, namely, credit-card-holder delinquencies and defaults, and further demonstrated that aggregated consumer-credit risk analytics may have important applications in forecasting systemic risks.

**Fraud detection:** Fraud detection involves monitoring the behavior of populations of users to estimate, detect, or avoid undesirable behavior. Undesirable behavior usually entails fraud, money laundering, cheating, and account defaulting. Outlier detection is the conventional technology for fraud detection. Bolton and David (2002) proposed an unsupervised credit card fraud detection method that incorporates other information (other than simply the amount spent) into the anomaly detection process. Neural networks can also be adopted in fraud detection. Aleskerov, Freisleben et al. (2002) presented a database mining system called CARDWATCH for credit card fraud detection. CARDWATCH uses a simple neural network to process current spending patterns and detect possible undesirable behavior. Rushin, Stancil et al. (2017) compared the performance of different methods in fraud detection and concluded that deep learning is the most accurate technique in predicting credit card fraud, followed by GBDT and LR.

**Bankruptcy prediction:** Bankruptcy prediction is the art of forecasting bankruptcy and various measures of financial distress of public firms. Bankruptcy prediction is relevant to creditors and investors in evaluating the likelihood that a firm may become bankrupt. Numerous accounting ratios that might indicate danger can be calculated, and other potential explanatory variables are available. Many studies on bankruptcy and insolvency prediction compared various approaches, modeling techniques, and individual models to ascertain whether any one technique is superior to its counterparts. For example, Min, Lee et al. (2006) applied SVM and the genetic algorithm (GA) to the problem of bankruptcy prediction. Kumar and Ravi (2007) presented a comprehensive review of research on statistical and intelligent techniques to solve the bankruptcy prediction problem faced by banks and firms. Olson, Delen et al. (2012) utilized various data mining tools to acquire bankruptcy data with the purpose of comparing accuracy and number of rules.

### 2.3 Financial identity authentication

Identity authentication is the key to ensuring financial information security. Identifying users through recognition, image recognition, voice print recognition, and OCR technology would significantly reduce checking costs and improve user experience. Identity authentication is a complete confirmation process used for the relationship between an entity and its identity in the network world. Government and private organizations use different biological characteristics to automate human recognition (Kedia and Monga 2017). Fig. 1 shows the technology landscape of biometric solutions[1]. In the banking industry, existing applications include identity verification, face withdrawals, and face recognition payments (Wang, Zhou et al. 2017). On September 1, Alipay's "face pay" system went online at KFC's KPRO restaurant. The system allows users to pay by brushing their faces and is the first commercial use of face recognition technology. It adopts EyePrint ID technology in payment scenarios and can thus overcome light, expression, makeup, age, and even cosmetic technical barriers based on cameras. The system can resist attacks. The new iPhone X uses Face ID to replace Touch ID for phone unlocking and payment certification. To further ensure financial security in the era of new technology, recognizing user identity and non-face hacks can be an important research area. SenseTime's unique face biometric detection technology can effectively prevent various non-face hacks and accurately determine the current validator for the user, rather than using other non-positive photos or videos, to ensure user information security.

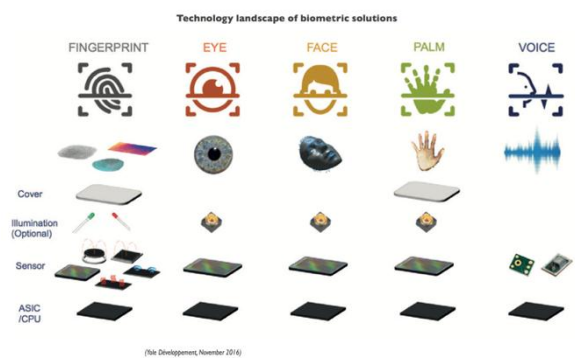

**Fig. 1 Technology landscape of biometric solutions***
* Source:  Yole Development, November, 2016

Many studies have been conducted on unconstrained face recognition using deep learning technology, such as DeepFace (Taigman, Yang et al. 2014) and DeepID (Sun, Wang et al. 2014). Sun, Wang et al. (2014) demonstrated that deep learning provides a powerful tool for undertaking the main challenges in face recognition, including reducing the internal differences and expanding the differences among people. Kuang, Huang et al. (2015) implemented automatic facial age estimation with deep CNN by fusing random forest and quadratic regression with local adjustment.

### 2.4 Smart financial consulting

Recent developments on speech recognition and natural language processing enable machine-to-human communication via interactive smart Q&A interfaces (e.g., chatbot systems such as Siri and Cortana) (Etzioni 2011), which enhance service experience while reducing costs. Generally, chatbot systems can analyze customers' goals and are highly responsive to customers with personalized advice or tailored answers, such as investment policies and portfolio strategies. The core tasks in designing a smart Q&A system can be roughly divided into five procedures: (1) speech recognition and synthesis, (2) named entity recognition and query intention understanding, (3) sentiment analysis and user profiling, (4) knowledge-based question answering (KB-QA), and (5) dialogue management and generation. In this study, we review state-of-the-art approaches on (4) and (5).

---

[1]   https://www.i-micronews.com/images/Flyers/MEMS/Yole_Sensors_for_Biometry_and_Recognition_2016_Report_Flyer.pdf

**KB-QA:** Existing approaches for KB-QA use learnable components to either transform the question into a structured KB query via semantic parsing (Berant, Chou et al. 2013, Yih, Chang et al. 2015) or embed questions and facts in a low-dimensional vector space and retrieve the answer by computing similarities (Bordes, Chopra et al. 2014, Dong, Wei et al. 2015). For example, Yih, Chang et al. (2015) initially defined a query graph that transforms semantic parsing into query graph generation with staged states and actions and then utilized deep CNN to perform semantic matching. A current trend in literature involves the utilization of memory networks (MemNNs) (Bordes, Usunier et al. 2015) and attention mechanisms (Andreas, Rohrbach et al. 2016, Zhang, Liu et al. 2016) to perform KB-QA. Bordes, Usunier et al. (2015) presented a MemNN-based QA system that stores facts from KBs via a memory component and performs generalization and inference in the memory. Zhang, Liu et al. (2016) employed an attention model to represent questions dynamically according to different answer aspects and integrated the global KB information into the representation of the answers to alleviate the out of vocabulary problem. Moreover, Andreas, Rohrbach et al. (2016) proposed a dynamic neural module network that uses a collection of compositional, attentional modules to leverage the best aspects of conventional logical forms and continuous representations.

**Dialogue management and generation:** Practical dialogue systems consist of a natural language understanding module, a natural language generation module, and a dialogue management module with state trackers and dialogue policy that tracks the state evolution and chooses an action given the current state. Traditional dialogue systems use a set of pre-programmed rules and thus cannot cope with non-stationary user behaviors (Li, Lipton et al. 2016). Hence, reinforcement learning (Graves, Wayne et al.), in which policies are learned automatically from experience and evolve from interactions with users, offers an appealing alternative (Dhingra, Li et al. 2016, Li, Monroe et al. 2017, Li, Chen et al. 2017). Dhingra, Li et al. (2016) presented an end-to-end RL-based dialogue agent called KB-infoBot to perform knowledge-based accessing, dialogue state tracking, and policy learning under a multi-turn setting. KB-infoBot showed great promise when applied to a goal-oriented task. Moreover, Li, Chen et al. (2017) presented a task-completion dialogue system to complete a task (e.g., movie ticket booking). The open domain dialogue generation problem was considered by Li, Monroe et al. (2017), who aimed to generate meaningful and coherent dialogue responses. An adversarial dialogue generation system was proposed in their work, and RL was applied to jointly train two sub-systems: a generator to produce response sequences and a discriminator to distinguish between the human-generated dialogues and the machine-generated ones.

### 2.5 Blockchain

Although the big data era has arrived, data are often mastered in different institutions. Data ownership, data security, and credit intermediaries thus become barriers to data sharing. A new generation of Internet technology is required to solve the problem of information decentralization. Hence, blockchain was introduced. The concept of blockchain was proposed by Nakamoto in 2008 (Nakamoto 2009). Blockchain, which underpins bitcoin, is a digital currency supported by cryptographic methods (a "cryptocurrency") and is a distributed, publicly available, and immutable ledger. In finance, blockchain has broad application prospects in digital currency, payment and settlement, intelligent contracts, and financial transactions. Typical applications include bitcoin, litecoin, and other electronic currencies, more secure and open distributed accounting systems, and payment and settlement systems. Consensus mechanisms and security guarantees are important components of blockchain technology.

**Consensus mechanism of blockchain:** Reaching a consensus efficiently in a distributed system and ensuring the security of transaction data are important research issues. Ethereum, the first Turing-complete decentralized smart contract system (Buterin 2014), uses proof of work (PoW) to ensure the consistency of distributed bookkeeping. However, PoW relies heavily on computing power and consumes extensive resources. Researchers have recently proposed various consensus mechanisms, such as proof of stake (PoS) and algorand. PoS (Reed 2014), also known as equity proof, changes the computing power of PoW to the system rights and interests. The larger the equity is, the greater the probability of becoming the next bookkeeper is. The algorand consensus algorithm (Micali 2016) was proposed by Professor Silvio Micali, a Turing Award winner, to solve the problem of having too many nodes in public chains.

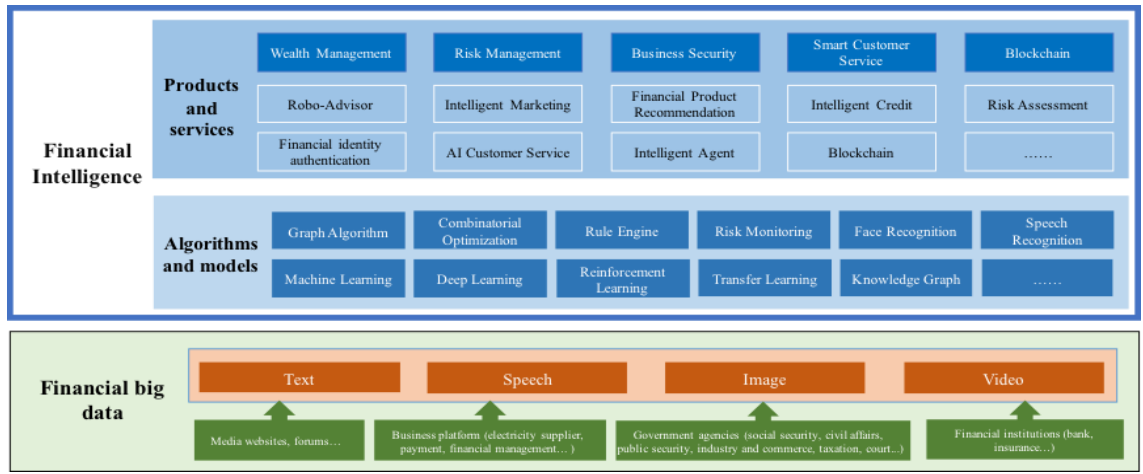

**Fig. 2 FinBrain framework: overall research structure**

**Security in blockchain:** Security issues have always been the focus of blockchain technology. The PoW mechanism faces 51% attacks. The PoS mechanism solves the problem of 51% attacks to a certain extent but introduces the nothing at stake (N@S) attack problem (Rosenfeld 2014). At present, no ideal solution is available for all types of attacks, and further research breakthroughs are needed.

## 3 FinBrain Framework and Open Issues

This section presents FinBrain, a research framework, and discusses several emerging issues.

### 3.1 FinBrain

Despite the research efforts devoted to financial intelligence, further progress is still needed to develop a more advanced technology that meets the requirements of inclusive finance. We envision that future finance based on AI 2.0 is a finance of AI, where AI agents will become more rational than people when making decisions, and a multi-agent system of AIs will be the core component of financial transactions.

Accordingly, we propose a research framework called FinBrain and further discuss several emerging open issues in Sec. 3.2. The overall research structure includes three main levels: (1) financial big data, (2) algorithms and models, and (3) products and services (Fig. 2). The financial big data level focuses on multi-source heterogeneous information processing, such as combining the structural financial report data and unstructured behavioral data of financial users via a knowledge graph. Through an analysis of different financial scenarios, we can enhance the perception capability for further prediction. The algorithm and model level aims to support different AI algorithms and models for financial intelligence, such as deep learning, reinforcement learning, and combinatorial optimization. Through a new generation of AI technology, we can further improve the decision-making capability and have the potential of constructing rational agents. The financial intelligence product and service level typically includes five major financial application scenarios, which are discussed in Sec. 2.

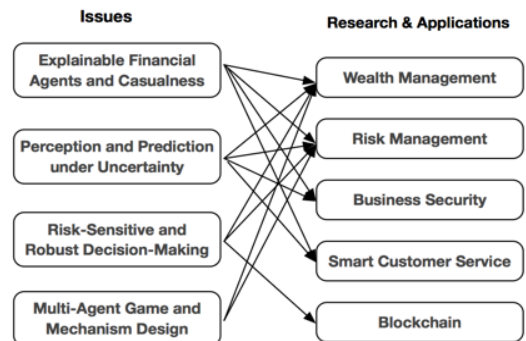

**Fig. 3 Mapping of issues, applications, and research topics**

### 3.2 Open Issues

To realize FinBrain in the era of AI 2.0, we must address crucial issues from the technical perspective,

including (1) explainable financial agents and causality, (2) perception and prediction under uncertainty, (3) risk-sensitive and robust decision making, and (4) multi-agent game and mechanism design. We believe that these open issues and future directions can lay the foundation for the development of AI 2.0 in the finance field. Fig. 3 shows the most common relationships among issues on the one hand and applications and research topics on the other hand. We describe these emerging issues in detail.

3.2.1 Explainable Financial Agents and Causality

A key component of financial agents is the ability to provide explanations for their decisions, predictions, or recommendations. However, current models usually make black-box predictions but rarely explain the process of decision making in a way that is meaningful to humans. We stress that explainable decisions are important to financial users. For example, investors require understanding before committing to decisions with inherent risks. In general, explainable agents should be able to identify the properties of the input (i.e., feature interactions) that are responsible for the particular decision and should be able to further capture the causal relations to answer counterfactual questions. The key to designing explainable agents is to provide causal inference to better understand the real-world environment and support interactive diagnostic analysis that faithfully replays a prediction against past perturbed inputs to measure feature importance. Module neural networks (Andreas, Rohrbach et al. 2016) may be a suitable choice for supporting explainable decisions. These networks utilize dynamic attention mechanisms to form predictive network layouts from a collection of composable modules.

3.2.2 Perception and Prediction under Uncertainty

Given the dynamic nature of financial environments (e.g., fluctuations in financial markets), perception and prediction agents that provide financial insights and reveal what will happen next should be able to process incomplete information and capture temporal correlation patterns under uncertainty. To satisfy these requirements, agents must form inner world models that represent comprehensive knowledge about the local environment by continuously perceiving multiple modalities and selectively integrating information from unstructured data sources. In particular, a memory mechanism (Bordes, Usunier et al. 2015, Graves, Wayne et al. 2016) with selective read/write heads and memory generalization operation can be employed to combine prior knowledge with temporal characteristics of sequences and thus allows knowledge evolution along with the dynamic changes in financial markets or user interests. Moreover, imagination-based reinforcement learning (Hamrick, Ballard et al. 2017) that learns to optimize a sequence of imagined internal actions over predictive world models before executing them in the real world may be a good choice for safe prediction under complex nonlinear dynamics. We highlight that these intelligent perception and prediction methods should be applied to important financial applications, such as market monitoring, risk management, and identity authentication, to significantly improve their intelligent services.

3.2.3 Risk-Sensitive and Robust Decision Making

In mission-critical applications, such as risk management and portfolio recommendation, an important requirement for financial agents is that they must be risk-sensitive and robust to uncertainty and errors when making decisions. In general, this issue should be considered from two perspectives. On the one hand, we can build source-specific noise models that automatically track data provenance and learn to distinguish the impact of different data sources with confidence intervals (Stoica, Song et al. 2017). On the other hand, we can construct risk-sensitive decision models that incorporate risk factors into loss functions and optimize under the worst-case scenario. For example, instead of the standard risk-neutral MDP based on expectation, Chow, Tamar et al. (2015) proposed a conditional-value-at-risk (CVaR) MDP that minimizes a risk-sensitive CVaR objective to perform robust decision making. To assess risks from intelligent adversaries (i.e., adversarial risks), such as competitors or terrorists, we can adopt the game-theoretic framework for adversarial learning to simulate the adversaries' strategies and enhance security. Counterterrorism modeling and GANs may be good choices in this case.

### 3.2.4 Multi-Agent Game and Mechanism Design

Financial markets involve assets, traders, online platforms, supply chains, and logistics, and they can be considered as multi-agent systems (MASs). Learning to play a multi-agent game is crucial in robust credit systems, supply chain management, dynamic asset pricing, and trading mechanism design. For example, to achieve dynamic pricing, we need to design an effective monetary incentive mechanism and capture the supply and demand dynamics in the real world. In E-commerce platforms, we can optimize different interests and explore the equilibrium state in the game of customers, merchants, and recommendation agents to balance platform revenues and user experiences. To this end, three challenges should be addressed. First, given scalability problems, a small number of agents is usually considered, which hinders the deployment in businesses that involve large-scale agents. Second, game-theoretic-based agents rely on the assumption of perfect rationality of individuals, which is difficult to apply in real-world scenarios. Lastly, cooperative actions and rewards must be quantified. Moreover, a profound discussion has been conducted on the future of economic AI, wherein AI agents engage in business transactions with other AI agents as well as with firms and people. Therefore, we must construct rational agents (called "machina economicus" in DC and MP [2015]) and design the rules of interaction for AI agents in this new economic system. For example, to elicit truthful reports and promote fair contributions, we can associate AI systems with a reputation to prevent the problems of moral hazard and adverse selection and further promote cooperativeness between AI systems when completing financial transactions.

## 4 Conclusion

With the aid of AI 2.0, the financial industry has changed in all directions, such as diverse sources of information collection, intellectualization of risk pricing models, standardization of the investment decision-making process, automation of customer interaction services, and other financial core fields. These changes are based on two paths: one is to improve efficiency and the other is to reduce costs.

In this article, we review extensive theoretical studies on and industrial applications of financial intelligence in the AI 2.0 era. First, we describe the concept of financial intelligence and elaborate on its position in the fintech field. Second, we describe several key applications. Finally, we propose a framework of research called FinBrain and summarize four open issues. By presenting the approaches, applications, and future directions in the financial intelligence field, we draw attention to state-of-the-art advancements and provide technical insights by discussing the challenges and research directions in these areas. We believe AI 2.0 is bound to move financial services toward the direction of "high efficiency and intelligence."

## 5 Acknowledgement

This work was supported in part by the National Natural Science Foundation of China (No.U1509221), the National Key Technology R&D Program(2015BAH07F01)，the Zhejiang Province key R&D program (No.2017C03044).